\documentclass[conference,letterpaper,10pt]{IEEEtran}
\IEEEoverridecommandlockouts 

\usepackage[T1]{fontenc}
\usepackage[english]{babel}
\usepackage{textcomp}
\usepackage[table, pdftex, dvipsnames]{xcolor}
\usepackage{xspace}
\usepackage{graphicx}
\usepackage{amsfonts}
\usepackage{amsmath}
\usepackage{amssymb}
\usepackage{booktabs}
\usepackage[acronym]{glossaries}
\usepackage[capitalize,noabbrev]{cleveref} 
\usepackage{cite}
\usepackage{tikz}
\usepackage{multirow}
\usepackage{multicol}
\usepackage{colortbl}
\usepackage{subfigure}
\usepackage{siunitx}
\usepackage{algpseudocode}
\usepackage{algorithm}
\usepackage{url}
\usepackage[utf8]{inputenc}
\usepackage[most]{tcolorbox}
\usepackage{acro}

\glsdisablehyper
\AtBeginDocument{}
\tcbset{
    colback=gray!5,
    colframe=gray!40,
    left=6pt,
    right=6pt,
    top=6pt,
    bottom=6pt,
    boxsep=0pt,
}

\makeatletter
\DeclareRobustCommand\onedot{\futurelet\@let@token\@onedot}
\def\@onedot{\ifx\@let@token.\else.\null\fi\xspace}
\def\eg{\emph{e.g}\onedot} 
\def\ie{\emph{i.e}\onedot}

\makeatother

\DeclareAcronym{ai}{short=AI, long= Artificial Intelligence}
\DeclareAcronym{dsp}{short=DSP, long= Digital Signal Processing}
\DeclareAcronym{sram}{short=SRAM, long= Static Random-Access Memory} 
\DeclareAcronym{sdram}{short=SDRAM, long= Synchronous Dynamic Random-Access Memory}
\DeclareAcronym{mac}{short=MAC, long= Multiply-Accumulate}

\begin{document}

\AddToHookNext{shipout/foreground}{%
\begin{tikzpicture}[overlay, remember picture]
    \node at ([yshift=-1cm]current page.north) {
        \normalsize\textcolor{gray}{This paper has been accepted for publication at the}
    };
    \node at ([yshift=-1.5cm]current page.north) {
        \normalsize\textcolor{gray}{8th International Conference on Artificial Intelligence Circuits and Systems (AICAS), Ha Long Bay, Vietnam, 2026}
    };
\end{tikzpicture}%
}

\title{
TinyGLASS: Real-Time Self-Supervised In-Sensor Anomaly Detection

\thanks{
The research was funded by the Swiss National Science Foundation (Grant 219943).
\textsuperscript{\textsection}P. Bonazzi and R. Sutter contributed equally to this work.
The source code and dataset are available at: https://github.com/ETH-PBL/TinyGLASS.}
}

\author{
\IEEEauthorblockN{Pietro Bonazzi\IEEEauthorrefmark{1}\textsuperscript{\textsection},
Rafael Sutter\IEEEauthorrefmark{1}\textsuperscript{\textsection},
Luigi Capogrosso\IEEEauthorrefmark{2},
Mischa Buob\IEEEauthorrefmark{3},
Michele Magno\IEEEauthorrefmark{1}\IEEEauthorrefmark{2}}
\IEEEauthorblockA{
\IEEEauthorrefmark{1}ETH Zurich, Zurich, Switzerland\\
\IEEEauthorrefmark{2}Interdisciplinary Transformation University of Austria, Linz, Austria\\
\IEEEauthorrefmark{3}Swiss Engineering Partners AG, Zurich, Switzerland}
}

\maketitle

\bstctlcite{IEEEexample:BSTcontrol}

\begin{abstract}
Anomaly detection plays a key role in industrial quality control, where defects must be identified despite the scarcity of labeled faulty samples.
Recent self-supervised approaches, such as GLASS, learn normal visual patterns using only defect-free data and have shown strong performance on industrial benchmarks.
However, their computational requirements limit their deployment on resource-constrained edge platforms, especially within in-sensor processing architectures.
This work introduces TinyGLASS, a lightweight adaptation of the GLASS framework designed for real-time edge and in-sensor anomaly detection.
The proposed architecture replaces the original WideResNet-50 backbone with a compact ResNet-18 and introduces deployment-based modifications that enable static graph tracing and INT8 quantization.
We evaluate the proposed approach on the Sony IMX500 intelligent vision sensor, exploiting the in-sensor processor using the Sony Model Compression Toolkit.
In addition to evaluating performance on the MVTec-AD benchmark, we investigate robustness to contaminated training data and introduce a custom industrial dataset, named MMS Dataset, for cross-device evaluation. 
Experimental results show that TinyGLASS achieves 8.6$\times{}$ parameter compression while maintaining competitive detection performance, reaching 94.2\% image-level AUROC on MVTec-AD and operating at 20 FPS within the 8 MB memory constraints of the IMX500 platform.
System profiling showcases low power consumption (4.0 mJ per inference), real-time end-to-end throughput (20 FPS), and high energy efficiency (470 GMAC/J).
Furthermore, the model demonstrates stable performance under moderate levels of training data contamination. 
\end{abstract}


\section{Introduction} \label{sec:intro}

In modern industrial manufacturing, visual inspection remains a crucial element of quality control.
Although automation has advanced significantly, many inspection tasks are still performed by humans or by systems that require meticulously labeled data for training.
This requirement poses a major limitation in real-world scenarios where labeled defect data are scarce, especially in settings characterized by a wide product variety and limited production volume, commonly referred to as \emph{high-mix-low-volume} environments.

Recent developments in machine learning, particularly in anomaly detection, offer promising alternatives by enabling systems to learn normal visual patterns using only defects-free data \cite{Hojjati2024, Li2025}.
These approaches can detect irregularities without extensive defect annotation, making them attractive for industrial use \cite{Liu2024}.

However, small and medium enterprises with limited infrastructure and budgets often rely on low-cost edge computing devices for real-time quality control on their production lines \cite{Antonini2023, Khan2025}.
This dependence creates a gap between the high performance of advanced anomaly detection models and their practical deployment in real manufacturing environments \cite{Barusco2025}.

\begin{figure}[t!]
    \centering
    \includegraphics[width=\linewidth]{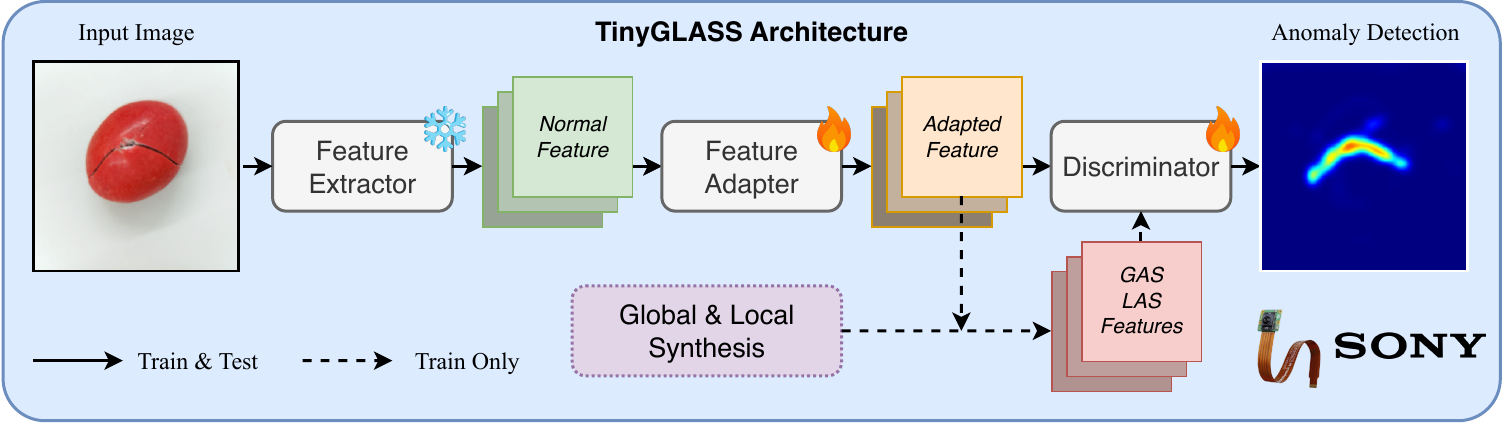}
    \caption{The proposed TinyGLASS architecture for end-to-end in-sensor anomaly detection running at \qty{20}{FPS} with only \qty{5.41}{MB} of total memory.}
    \label{fig:TinyGLASS}
    \vspace{-0.5cm}
\end{figure}
To address this challenge, we present TinyGLASS (shown in \cref{fig:TinyGLASS}), a novel self-supervised anomaly detection model designed for in-sensor computing \cite{Pau2025, Capogrosso2026}.
Our main contributions are threefold.

First, we introduce TinyGLASS, a lightweight adaptation of the GLASS framework that replaces the original WideResNet-50 \cite{Zagoruyko2016} backbone with a compact ResNet-18 \cite{He2016} and incorporates deployment-aware modifications (static graph tracing and INT8 quantization) compatible with Sony's Model Compression Toolkit \cite{SonyMTC}.
Second, TinyGLASS achieves 8.6$\times{}$ parameter compression while providing competitive performance (\qty{94.2}{\percent} image-level AUROC on MVTec-AD \cite{Bergmann2019}) and runs at \qty{20}{FPS} using only \qty{5.41}{MB} on the new Sony IMX500 intelligent vision sensor \cite{IMX500}, representing the first demonstration of real-time in-sensor visual anomaly detection.
Third, we thoroughly analyze robustness to contaminated training data and introduce a new custom industrial dataset (``MMS Dataset'') captured with both a high-resolution microscope and the IMX500 sensor, enabling cross-device evaluation.
\section{Related Work} \label{sec:related}

\subsection{Industrial Anomaly Detection}

\emph{\textbf{Reconstruction-based methods}} detect anomalies by analyzing the residual image before and after reconstruction under the assumption that a model trained only on normal data will fail to accurately reconstruct abnormal regions.
Early approaches used Autoencoders (AEs) and Variational Autoencoders (VAEs) \cite{Gong2019}.
Other methods, such as those proposed in \cite{Zavrtanik2021a, Ristea2022}, frame anomaly detection as an inpainting problem in which image patches are randomly masked.
However, they are heavily dependent on the quality of the reconstructed image.
If anomalies share common patterns (\eg{}, local edges) with normal data, or if the decoder is ``too strong', defects may be reconstructed accurately, leading to small residual errors that fail to distinguish between normal and abnormal areas \cite{Zavrtanik2021a}.

\emph{\textbf{Embedding-based methods}} utilize pre-trained networks to extract features, which are then used to learn normality.
Memory bank methods, such as PatchCore \cite{Roth2022} and PNI \cite{Bae2023}, represent normal features in an archive and detect anomalies using metric learning.
Similarly, one-class classification methods, such as PANDA \cite{Reiss2021}, map image features into a compact latent space where normal data are tightly clustered, allowing anomalies to be identified as points that fall outside this region.
Other distribution-based approaches, such as PaDiM \cite{Defard2021}, model the normal state using Gaussian distributions at each pixel location.
To better handle complex data, Normalizing Flow methods, such as FastFlow \cite{Yu2021}, DifferNet \cite{Rudolph2021}, and CFLOW-AD \cite{Gudovskiy2022}, learn to transform normal feature distributions into a standard probability space.
However, either computing the inverse of the covariance \cite{Defard2021} or searching for the nearest neighbor in the memory bank \cite{Roth2022} limits their practical implementation and real-time performance on ultra-low-power devices.

\emph{\textbf{Synthesis-based methods}} view the synthesis of anomalies as a form of data augmentation, creating ``fake'' defects on normal images to train a model to recognize what a deviation looks like \cite{Capogrosso2024a, Girella2024}.
CutPaste \cite{Li2021} is the pioneering work in this area, generating anomalies by cutting a patch from one image and pasting it into another.
DRAEM \cite{Zavrtanik2021b} improved this by using Perlin noise to simulate more realistic textures and training a network to locate and repair defects.
Recently, SimpleNet \cite{Liu2023} has achieved high accuracy using a simple discriminator to separate normal features from synthetic noise.
GLASS \cite{Chen2024} was proposed on top of SimpleNet, combining image-space corruptions with feature-space perturbations that improve weak defect detection.

\subsection{End-To-End Edge Vision}
The design and deployment of neural networks for edge devices have attracted increasing attention in recent years \cite{Capogrosso2024b}.
Approaches typically balance model accuracy, computational efficiency, and hardware constraints.
Consequently, they often rely on hardware-friendly operations (such as convolution micro-factorizations and lightweight network designs) that can be accelerated on embedded platforms.

In parallel, several works have evaluated deep learning models on a range of edge and embedded systems \cite{Giordano2022, Bonazzi2023}.
For example, MCUNet \cite{Lin2020} demonstrated end-to-end image classification on microcontrollers at $\approx$\qty{10}{FPS}, and TinyissimoYOLO \cite{Moosmann2023} demonstrated an end-to-end object detection latency of \qty{56}{ms} and a throughput of \qty{18}{FPS} \cite{Moosmann2024}.

More recently, in-sensor Artificial Intelligence (AI) platforms such as Sony IMX500 \cite{Eki2021, IMX500} have enabled vision models to run directly on the image sensor, reducing system latency and energy consumption \cite{Bonazzi2023}.
The usual applications of this platform range from general-purpose image segmentation \cite{Bonazzi2025} to health monitoring \cite{Tong2024, Bonazzi2024} and smart cities \cite{Cui2024}.

Despite these advances, end-to-end visual anomaly detection has not yet been demonstrated on in-sensor low-power edge platforms. 
\section{Methodology} \label{sec:methodology}

\subsection{Hardware Setup}
The target deployment platform is the Sony IMX500 intelligent vision sensor \cite{Eki2021, IMX500}, a \qty{12.3}{MP} CMOS sensor ($\approx4056\times{}3040$ effective pixels, \qty{1.55}{\micro\meter} pixel size) that integrates an Image Signal Processor (ISP), a neural network accelerator, and on-chip AI inference capabilities.
This enables the execution of quantized deep neural networks directly on the sensor, producing metadata or anomaly scores rather than full images, thus reducing bandwidth, latency, and energy consumption in edge applications \cite{Bonazzi2023, Capogrosso2026}.
In recent years, the IMX500 has been integrated via the Raspberry Pi AI Camera \cite{AICamera}, which connects to a Raspberry Pi 5 via the standard MIPI CSI-2 interface.
Network inference produces patch-level anomaly heatmaps directly from the sensor, and, if required, final image-level decisions are aggregated on the host (Raspberry Pi).
This setup enables in-sensor real-time anomaly detection.

\subsection{TinyGLASS Architecture}
TinyGLASS is an efficient anomaly detection architecture adapted from GLASS \cite{Chen2024} for deployment on resource-constrained embedded platforms, such as the Sony IMX500 intelligent vision sensor.

Although some components of the original GLASS pipeline remain effective on GPU-based platforms, they are not compatible with the constraints imposed by embedded proxy tracking and quantization frameworks used for low-power processing.
The adaptations proposed here preserve the core anomaly detection capabilities while enabling efficient real-time execution on resource-constrained hardware.

First, the original GLASS architecture is based on a WideResNet-50 \cite{Zagoruyko2016} backbone that produces 1536-dimensional embeddings from concatenated level-2 and level-3 features.
To reduce memory footprint and computational demands while preserving the multi-scale feature fusion strategy, we replace this backbone with ResNet-18 \cite{He2016}, producing 384-dimensional concatenated embeddings from the corresponding layers.
Intermediate feature extraction, previously performed via PyTorch hooks, is replaced by a modified forward pass that directly returns only the features from the required level-2 and level-3 layers.
This change enables the pruning of unused components and ensures compatibility with the static graph tracing required for hardware-aware quantization \cite{SonyMTC}. 

Dynamic tensor reshaping and unsupported operations are also eliminated to satisfy strict operator constraints during tracing, which replaces tensors with symbolic proxies and invokes the \texttt{forward} method under static shapes.
In addition, the original input path of the discriminator included a \texttt{flatten} operation followed by linear and convolutional layers, resulting in dynamic shapes that violated these constraints.
We resolve this by maintaining 4D tensor representations end-to-end, enabling standard \texttt{Conv2D} operations throughout.
Furthermore, the tuples returned between modules are replaced with single-tensor outputs; specifically, the interpolated features from layers 2 and 3 of the backbone are concatenated into a unified tensor.

The final network output is a single tensor of patch-level anomaly scores (\ie{}, a heatmap), while image-level scores are computed on the host side.
For in-sensor evaluation on the IMX500 \cite{Eki2021, IMX500}, compression is performed using Sony’s Model Compression Toolkit \cite{SonyMTC}, with both weights and activations quantized to INT8.

In terms of model complexity, the ResNet-18 backbone used in TinyGLASS contains 14.6~M parameters when fully instantiated.
However, only layers up to \texttt{layer3} are executed during inference, resulting in an effective backbone size of 2.78~M parameters.
The discriminator head, identical to that of GLASS, adds an additional 67.5~K parameters.
Including the remaining intermediate components (\eg{}, PatchMaker), the total number of parameters actively used at inference is $\approx$2.9~M.
Compared to 24.9~M for the corresponding GLASS configuration based on WideResNet-50, this produces an effective compression ratio of 8.6$\times$.

\subsection{Training Objective \& Implementation Details}
As illustrated in \cref{fig:TinyGLASS}, TinyGLASS leverages two complementary feature representations, namely the Global Anomaly Score (GAS) and the Local Anomaly Score (LAS) proposed in \cite{Chen2024}.
GAS captures coarse, image-level deviations by aggregating global contextual information, while LAS focuses on fine-grained, spatially localized discrepancies that are critical for detecting small or subtle anomalies.

The training objective combines two complementary loss terms to effectively distinguish anomalous samples from normal ones:
\begin{equation}
    \mathcal{L}_{\text{total}} = \mathcal{L}_{\text{BCE}} + \mathcal{L}_{\text{focal}}\;,
\end{equation}
where $\mathcal{L}_{\text{BCE}}$ is the Binary Cross-Entropy loss \cite{Prince2023} used for discriminator training, and $\mathcal{L}_{\text{focal}}$ is the Focal Loss \cite{Lin2017} applied to handle the class imbalance between normal and defective regions.

Our training protocol follows the standard unsupervised anomaly detection framework using only good samples during training, unless otherwise stated.
Patch-level anomaly scores are aggregated to produce final image-level predictions.
Data augmentation includes random rotations, translations, color jittering, and horizontal/vertical flips, applied with probability 0.5 to increase the model's robustness.

The model is trained using the AdamW optimizer \cite{Loshchilov2017} with a learning rate of $10^{-4}$ for the feature extraction components and $2\times{}10^{-4}$ for the discriminator.
We train on three NVIDIA RTX A6000 for a maximum of 200 epochs, with a batch size of 8.
The best model is selected based on image-level AUROC (I-AUROC) performance.

\subsection{Datasets}
We evaluate the proposed TinyGLASS model on two complementary datasets: \emph{i)} our newly introduced \emph{MMS Dataset}, which captures real-world industrial micro-component defects using both a high-resolution microscope and the target Sony IMX500 sensor, enabling direct assessment of domain transfer and edge-deployment feasibility; and \emph{ii)} the widely used MVTec-AD benchmark \cite{Bergmann2019}, a standard industrial anomaly detection dataset that allows comparison with prior methods and controlled experiments on training-set contamination to simulate realistic imperfect data collection scenarios.

\emph{\textbf{MMS.}}
To evaluate domain transfer and real-world applicability, we introduce the custom MMS Dataset, collected under controlled laboratory conditions.
The images were captured using a stereo zoom microscope (RYF AG) with a trinocular head mounted on an articulated boom arm.
A high-resolution camera stream was monitored in real time, enabling precise adjustment of exposure, gain, contrast, and measurement overlays.
The samples were recorded with fixed magnification and controlled illumination to reduce acquisition variance. To further study domain shift, a subset of images was also captured using the Sony IMX500 intelligent vision sensor \cite{Eki2021, IMX500}, which allows future cross-device evaluation and facilitates experiments targeting edge deployment.
The final dataset comprises four classes, \ie{}, \emph{crack-hole}, \emph{scratch}, \emph{half}, and \emph{normal}, covering structural and surface-level anomalies.

\begin{figure}[t!]
    \centering
    \begin{minipage}{0.33\linewidth}
        \centering
        \includegraphics[height=0.22\textheight]{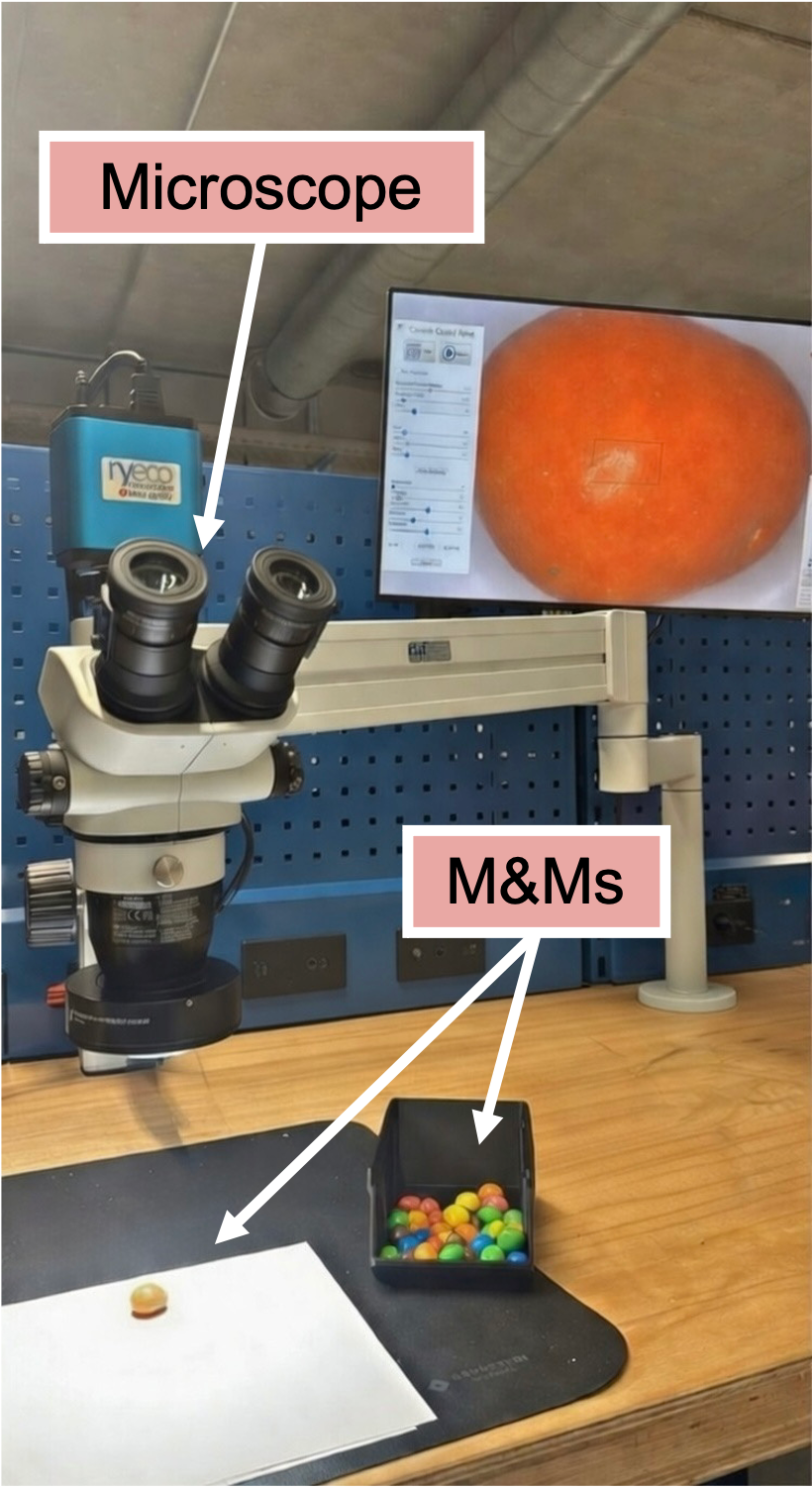}\par
        \vspace{0.4em}
        \small (a) Data Collection 
    \end{minipage}\hfill
    \begin{minipage}{0.63\linewidth}
        \centering
        \begin{tabular}{cc}
            \includegraphics[width=0.4\linewidth]{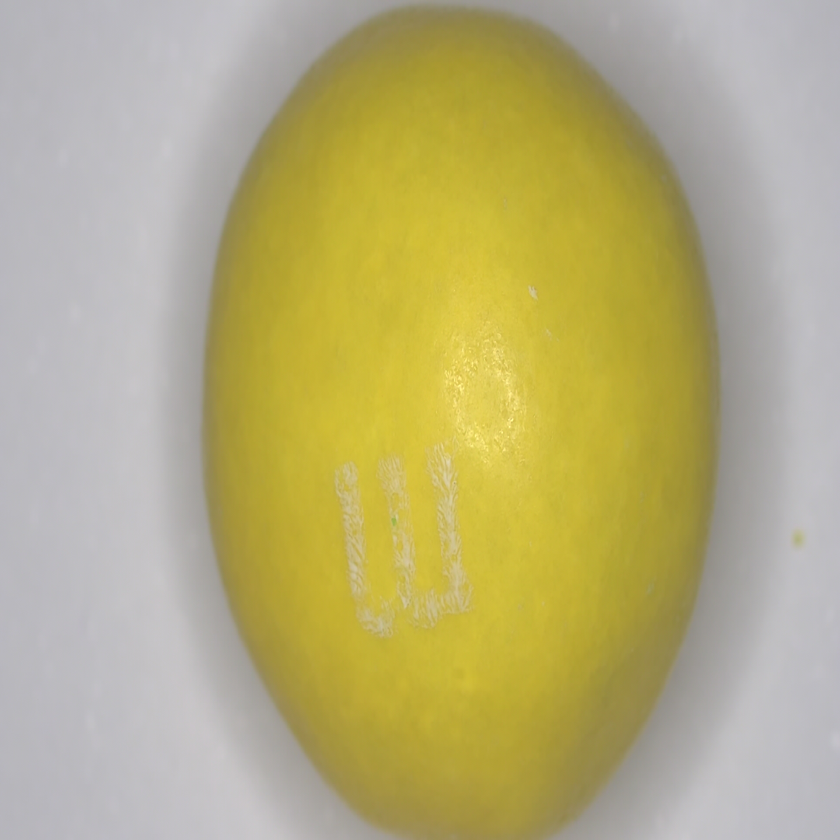} &
            \includegraphics[width=0.4\linewidth]{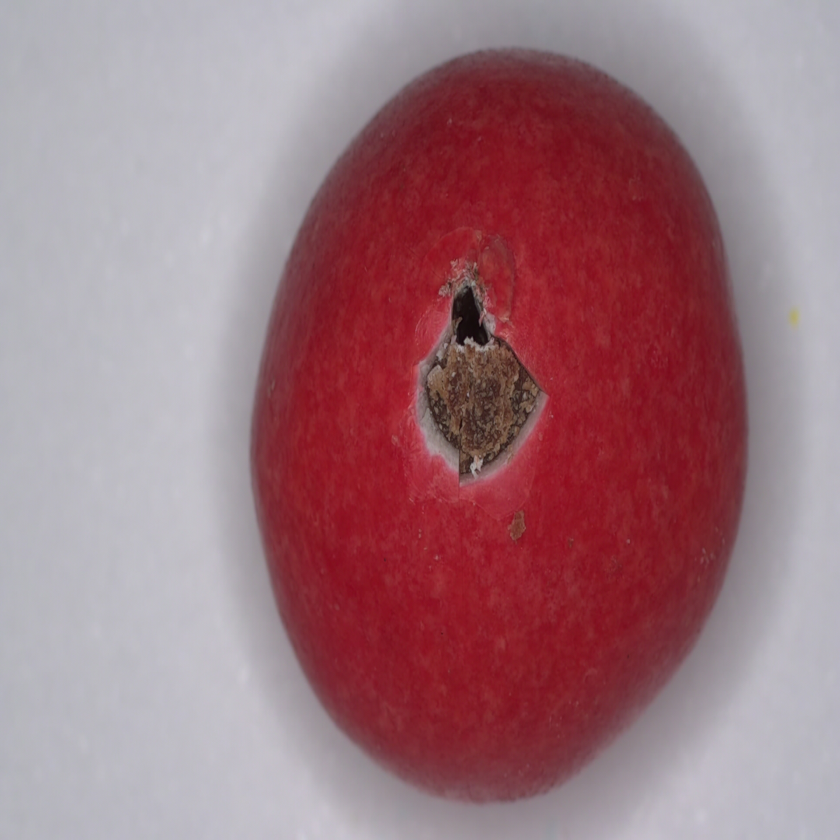} \\
            (b) Normal & (c) Crack-hole \\[0.4em]
            \includegraphics[width=0.4\linewidth]{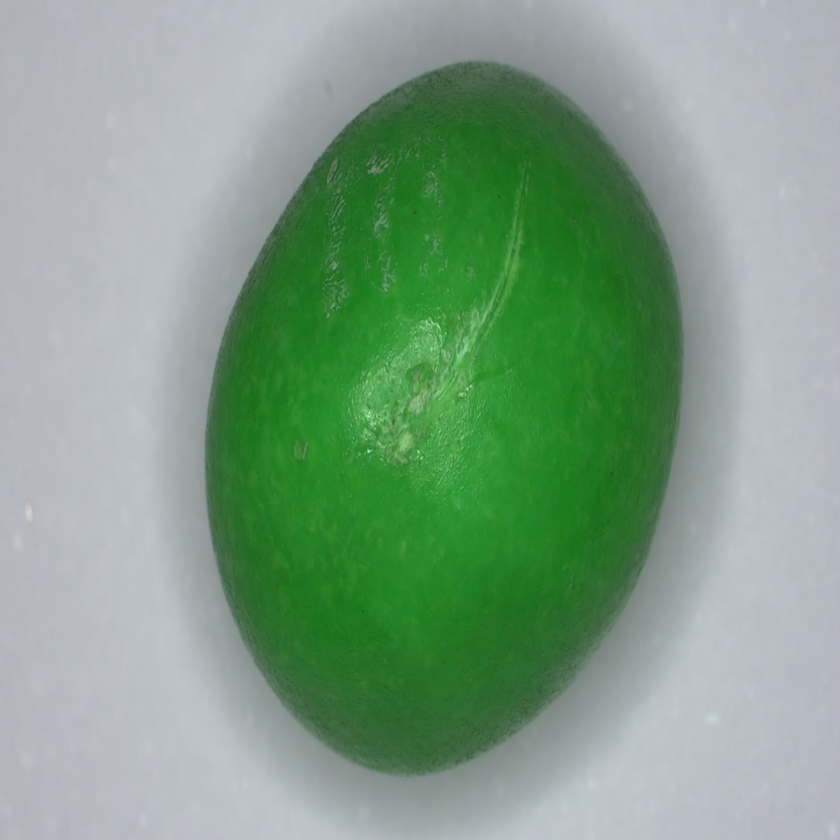} &
            \includegraphics[width=0.4\linewidth]{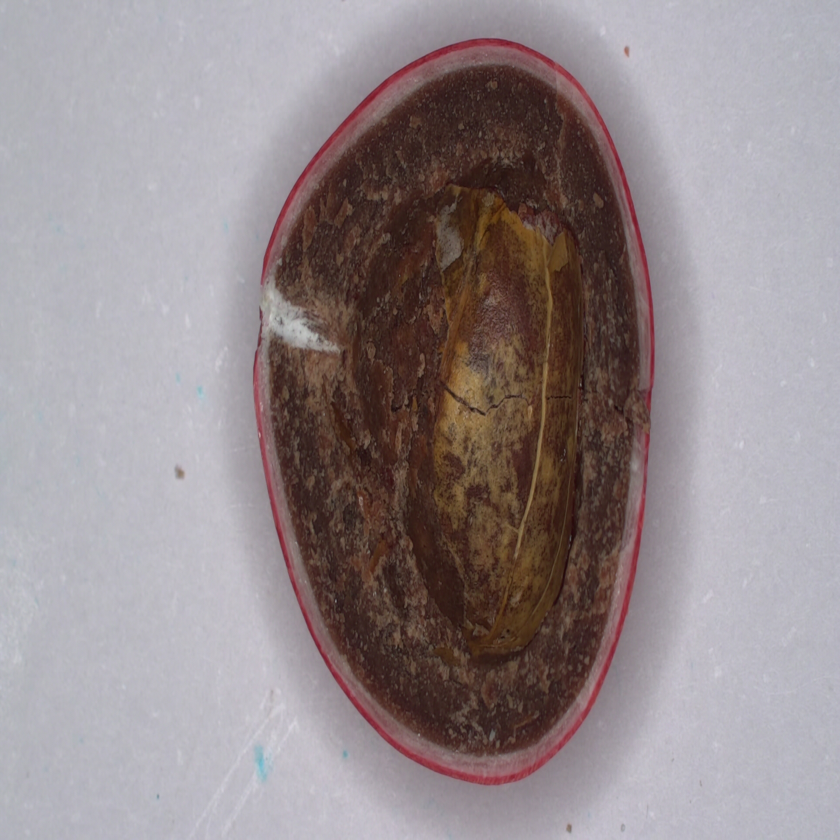} \\
            (d) Scratch & (e) Half
        \end{tabular}
    \end{minipage}
    \caption{MMS Dataset. Left: stereo microscope acquisition setup. Right: representative samples of the four classes.}
    \label{fig:mms_overview}
\end{figure}

\begin{table}[t!]
    \centering
    \small 
    \caption{Per-class sample distribution in the custom MMS Dataset across the two camera setups.}
    \label{tab:mms_dataset}
    \begin{tabular}{l c c c c}
    \toprule
    \textbf{Camera}       & \multicolumn{2}{c}{\textbf{Microscope}} 
                          & \multicolumn{2}{c}{\textbf{IMX500}} \\
    \cmidrule(lr){2-3} \cmidrule(lr){4-5}
    \textbf{Category}     & \textbf{Train} & \textbf{Test+Val} & \textbf{Train} & \textbf{Test+Val} \\
    \midrule
    Normal / Good         & 166            & 42                & 184            & 55 \\
    Crack-hole            & 0              & 18                & 0              & 28 \\
    Scratch               & 0              & 39                & 0              & 42 \\
    Half                  & 0              & 14                & 0              & 14 \\
    \midrule
    Defective (all types) & \textbf{0}     & \textbf{71}       & \textbf{0}     & \textbf{84} \\
    \midrule
    \textbf{Total}        & \textbf{166}   & \textbf{113}      & \textbf{184}   & \textbf{139} \\
    \bottomrule
    \end{tabular}
    \vspace{-0.2cm}
\end{table}
In \cref{tab:mms_dataset}, we provide statistics for two camera setups: the high-resolution microscope camera used for primary experiments and the IMX500 camera for additional domain-transfer evaluation.
\cref{fig:mms_overview} shows representative examples and illustrates the acquisition setup.

\emph{\textbf{MVTec-AD.}}
The MVTec-AD benchmark \cite{Bergmann2019} provides defect-free training images and pixel-annotated anomalous test samples across 15 industrial categories.
We resized the input images to $256\times{}256$ pixels and normalized them using ImageNet statistics \cite{Deng2009} before processing them by the network.
To simulate realistic industrial conditions where normal-only training data may be contaminated, we inject anomalous samples into the training set for the carpet category.
We vary the contamination ratio between \qty{5}{\percent} and \qty{30}{\percent}, sampling defective images from the held-out subsets.
This setting enables controlled analysis of robustness under imperfect supervision.
\section{Experimental Results} \label{sec:experiments}

We evaluate TinyGLASS in terms of detection accuracy and deployment efficiency.
First, we compare TinyGLASS with the original GLASS model on the MVTec-AD benchmark.
Next, we analyze the robustness to training-set contamination.
Finally, we report system-level performance of the model deployed on the target hardware platform.

\begin{table}[t!]
    \centering
    \small
    \caption{Comparison of GLASS and TinyGLASS on MVTec-AD (mean results).
    TinyGLASS achieves $8.6\times$ parameter compression with a \qty{4.9}{\percent} I-AUROC drop.}
    \label{tab:tinyglass_vs_glass}
    \begin{tabular}{l l l c c}
        \toprule
        \textbf{Model} & \textbf{Format} & \textbf{Params} & \multicolumn{2}{c}{\textbf{AUROC (\%)}} \\
        \cmidrule(lr){4-5}
        &  &  & \textbf{Image} & \textbf{Pixel} \\
        \midrule
        GLASS \cite{Chen2024} & float-32 & 24.9~M & 99.1 & 98.3 \\       
        TinyGLASS (ours)      & float-32 & 2.9~M  & 94.6 & 92.9 \\        
        TinyGLASS (ours)      & int-8    &  -            & 94.2 & 90.9 \\  
        \bottomrule
    \end{tabular}
\end{table}

\cref{tab:tinyglass_vs_glass} reports the mean performance on MVTec-AD.
TinyGLASS achieves \qty{94.2}{\percent} I-AUROC and \qty{90.9}{\percent} pixel-level AUROC (P-AUROC), corresponding to drops of \qty{4.9}{\percent} and \qty{7.4}{\percent} compared to GLASS.
On the custom MMS Dataset, the model reaches \qty{88.9}{\percent} I-AUROC when trained and tested on the microscope images.

\begin{figure}[t!]
    \centering
    \includegraphics[width=.95\linewidth]{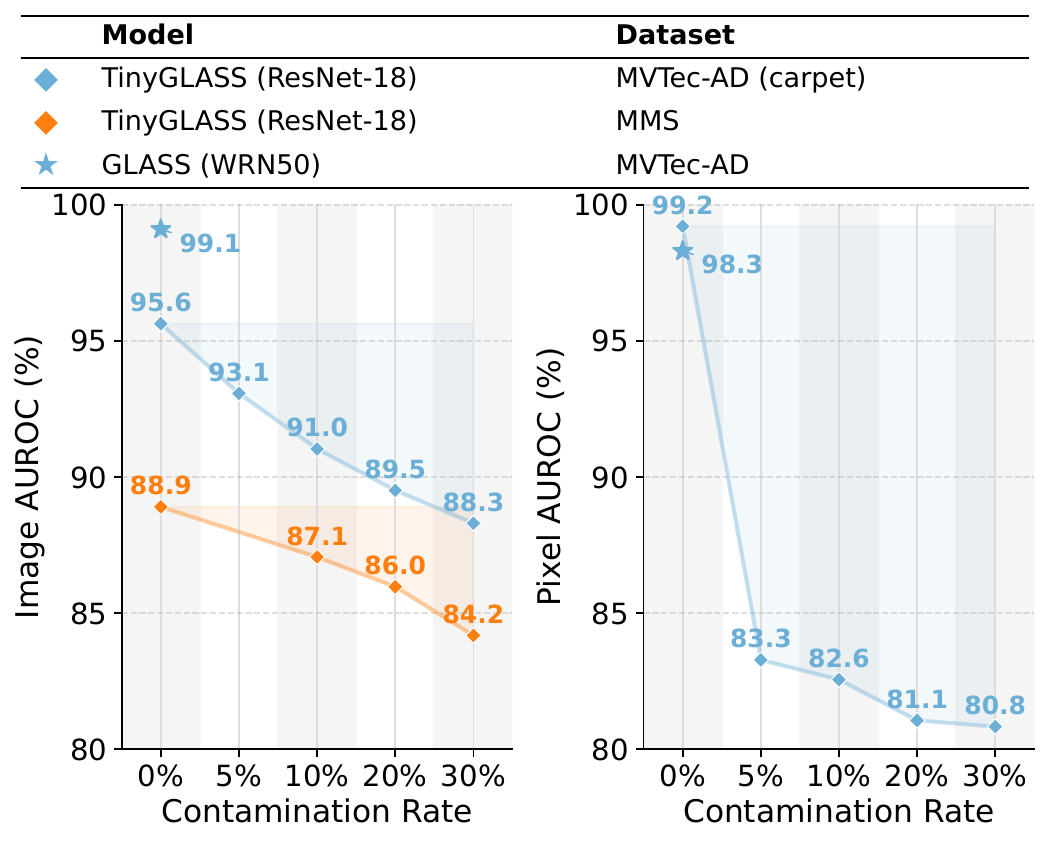}
    \caption{Effect of training-set contamination on TinyGLASS (ResNet-18) evaluated on MVTec-AD (\textit{carpet}) and MMS.
    I-AUROC (left) and P-AUROC (right) are shown for contamination rates ranging from \qty{0}{\percent} to \qty{30}{\percent}.
    GLASS (WideResNet-50) on full MVTec-AD is shown as a reference at \qty{99.2}{\percent}.}
    \label{fig:contamination}
    \vspace{-0.2cm}
\end{figure}

To evaluate robustness to training-set contamination, anomalous samples are injected into the training set at rates of \qty{0}{\percent}, \qty{5}{\percent}, \qty{10}{\percent}, \qty{20}{\percent}, and \qty{30}{\percent}.
I-AUROC and P-AUROC are reported in \cref{fig:contamination}.
On MVTec-AD (only \textit{carpet}), I-AUROC decreases from \qty{95.6}{\percent} at \qty{0}{\percent} contamination to \qty{88.3}{\percent} at \qty{30}{\percent}.
On MMS, I-AUROC decreases from \qty{88.9}{\percent} to \qty{84.2}{\percent} in the same range.
P-AUROC drops from \qty{99.2}{\percent} to \qty{80.8}{\percent} on MVTec-AD (\textit{carpet}) over the same contamination range.

Finally, we evaluate the TinyGLASS system deployed on the target edge platform.
In terms of runtime performance on the Sony IMX500, the system achieves $\approx$\qty{20}{FPS} during inference.
Each inference requires only 1.88~G Multiply-Accumulate Operations (GMACs) for $256\times{}256$ input images (1.844~G in the backbone, 0.016~G in PatchMaker, and 0.017~G in the discriminator).
This corresponds to an effective throughput of \qty{37.6}{GMAC/s}, an efficiency of \qty{470}{GMAC/J}, and an energy consumption of \qty{4.0}{mJ} per inference, among the lowest reported for in-sensor anomaly detection models.
\section{Conclusion \& Future Work} \label{sec:conclusions}

This work presents TinyGLASS, a lightweight self-supervised anomaly detection framework designed for deployment on resource-constrained edge vision in-sensor processors.
Experimental results demonstrate that TinyGLASS achieves substantial model compression while maintaining competitive detection performance.
Specifically, it reduces the parameter count by 8.6$\times{}$, while still achieving \qty{94.2}{\percent} I-AUROC.
System-level evaluation shows that the quantized model fits within the \qty{8}{MB} memory constraint of the target platform and achieves real-time inference at approximately \qty{20}{FPS} on the new Sony IMX500.
In addition, we evaluated robustness to contaminated training data and introduced a custom dataset for domain-transfer analysis, showing that TinyGLASS maintains stable performance under moderate levels of label misclassification.
Future work could explore further improvements in model robustness and efficiency, including lightweight backbone architectures \cite{Iandola2017, Howard2019, Tan2019}, broader validation across industrial inspection scenarios \cite{Zou2022}, and the investigation of explainable anomaly detection methods that provide interpretable localization and defect characterization.

\bibliographystyle{IEEEtran}
\bibliography{bibliography}

\end{document}